\pdfoutput=1

\documentclass[11pt]{article}

\usepackage[preprint]{acl}

\usepackage{times}
\usepackage{latexsym}

\usepackage[T1]{fontenc}

\usepackage[utf8]{inputenc}

\usepackage{microtype}

\usepackage{inconsolata}

\usepackage{graphicx}
\usepackage{pifont}
\usepackage{multirow}
\usepackage{makecell}
\usepackage{amsmath}
\usepackage{xspace}
\usepackage{soul}
\usepackage{tabularray}
\UseTblrLibrary{booktabs}

\usepackage{listings}
\usepackage{xcolor}
\lstdefinestyle{promptstyle}{
    basicstyle=\small\ttfamily,
    breaklines=true,
    frame=single,
    captionpos=b,
    keepspaces=true,
    numbers=none,
    showspaces=false,
    showstringspaces=false,
    showtabs=false,
    tabsize=2,
    backgroundcolor=\color{gray!10},
    columns=flexible,
    breakatwhitespace=true,
    postbreak=\mbox{\textcolor{red}{$\hookrightarrow$}\space},
}

\newcommand{\datasetname}{\texttt{OPeRA}\xspace}
\newcommand{\pluginname}{\texttt{ShoppingFlow}\xspace}

\newcommand{\checkmark}{\ding{51}}
\newcommand{\cross}{\ding{55}}

\title{\datasetname: A Dataset of \underline{O}bservation, \underline{Pe}rsona, \underline{R}ationale, and \underline{A}ction for Evaluating LLMs on Human Online Shopping Behavior Simulation}

\author{
 \textbf{Ziyi Wang\textsuperscript{1}},
 \textbf{Yuxuan Lu\textsuperscript{1}},
 \textbf{Wenbo Li\textsuperscript{2}},
 \textbf{Amirali Amini\textsuperscript{1}},
\\
 \textbf{Bo Sun\textsuperscript{1}},
 \textbf{Yakov Bart\textsuperscript{1}},
 \textbf{Weimin Lyu\textsuperscript{3}},
 \textbf{Jiri Gesi \textsuperscript{4}},
\\
 \textbf{Tian Wang\textsuperscript{4}}
 \textbf{Jing Huang\textsuperscript{4}},
 \textbf{Yu Su\textsuperscript{5}},
 \textbf{Upol Ehsan\textsuperscript{1}},
 \\
 \textbf{Malihe Alikhani\textsuperscript{1}},
 \textbf{Toby Jia-Jun Li\textsuperscript{6}},
 \textbf{Lydia Chilton\textsuperscript{7}},
 \textbf{Dakuo Wang\textsuperscript{1}},
\\
\\
 \textsuperscript{1}Northeastern University,
 \textsuperscript{2}University of Southern California,
 \textsuperscript{3}Stony Brook University,\\
 \textsuperscript{4}Independent Researcher,
 \textsuperscript{5}Ohio State University,
 \textsuperscript{6}University of Notre Dame,
 \textsuperscript{7}Columbia University
\\
 \small{
   \textbf{Correspondence:} \href{mailto:wang.ziyi19@northeastern.edu}{wang.ziyi19@northeastern.edu}, \href{mailto:d.wang@northeastern.edu}{d.wang@northeastern.edu}}
}

\begin{document}
\maketitle
\begin{abstract}
\textbf{Can large language models (LLMs) accurately simulate the next web action of a specific user?}
While LLMs have shown promising capabilities in generating ``believable'' human behaviors, evaluating their ability to mimic real user behaviors remains an open challenge, largely due to the lack of high-quality, publicly available datasets that capture both the observable actions and the internal reasoning of an actual human user.
To address this gap, we introduce \datasetname, a novel dataset of \textbf{O}bservation, \textbf{Pe}rsona, \textbf{R}ationale, and \textbf{A}ction collected from real human participants during online shopping sessions. 
\datasetname is the first public dataset that comprehensively captures: user personas, browser observations, fine-grained web actions, and self-reported just-in-time rationales.
We developed both an online questionnaire and a custom browser plugin to gather this dataset with high fidelity. 
Using \datasetname, we establish the first benchmark to evaluate how well current LLMs can predict a specific user's next action and rationale with a given persona and <observation, action, rationale> history.
This dataset lays the groundwork for future research into LLM agents that aim to act as personalized digital twins for human \footnote{https://huggingface.co/datasets/NEU-HAI/OPeRA}.

\end{abstract}

\section{Introduction}
\begin{figure}[t]
    \centering
    \includegraphics[width=\linewidth]{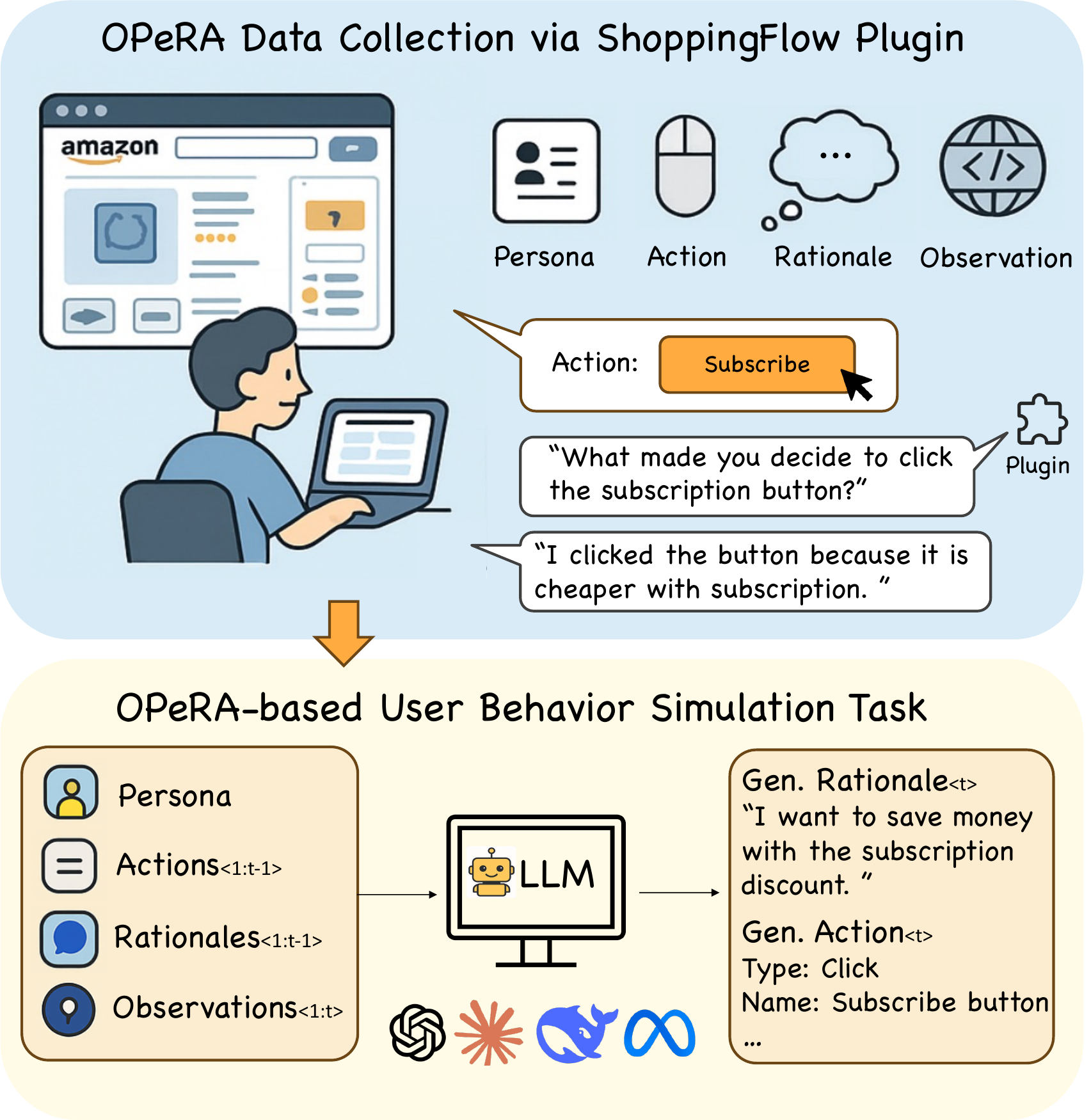}
    \caption{We developed \pluginname plugin (Figure~\ref{fig:pipeline}) to collect user shopping behavior over a four-week period, resulting in \datasetname-full dataset. 
    This dataset comprises 692 sessions from 51 unique users, containing 28,904 real-user <action, observation> pairs and 604 user-annotated rationales (Figure~\ref{fig:data-overview}). After postprocessing, we obtained \datasetname-filtered, which includes 527 sessions, 5,856 <action, observation> pairs, and 207 rationales. 
    We then benchmarked four LLMs' performance on user next action prediction task, results shown in Table~\ref{tab:action_predict_filtered}).}
    \label{fig:firstpage}
\end{figure}

Large language model (LLM) agents have exhibited impressive performance across diverse tasks, including planning, reasoning, and acting in web-based environments~\cite{xie2024travelplanner,yao2023react,jin2025search}.
A promising frontier in this area is human behavior simulation, where LLM agents generate user-like action sequences on digital platforms~\cite{chen2025towards}.
These agents (i.e., role-playing agents) are increasingly used in applications such as UI/UX testing~\cite{lu2025uxagentllmagentbasedusability}, social science research~\cite{parkGenerativeAgentSimulations2024}, accessibility testing~\cite{taebAXNavReplayingAccessibility2024}, and personal digital assistants~\cite{openai2025operator}.
Yet, while these agents can generate believable human behavior, the more critical question remains: can they generate behavior that accurately aligns with real human?

Despite progress in agent-based behavior simulation~\cite{park2023generative,parkGenerativeAgentSimulations2024}, current works still have limitations. First, most existing evaluations focus on aggregate outcomes (e.g., survey responses or end-task completions). These methods overlook the step-wise rationale and actions that underlie user behavior patterns~\cite{chen2025towards}.
For example, \citet{parkGenerativeAgentSimulations2024} compared LLM agents' survey results with real humans by replicating various social science studies. 
Furthermore, many role-play agents rely solely on prompting without grounding in real human data training, which limits their accuracy and personalization.
Although some recent efforts incorporate user behavior data via fine-tuning~\cite{lu2025believabilityaccuratehumanbehavior}, these datasets are often proprietary or lack critical detail, such as the reasoning or persona behind user actions.
 
Current open-source datasets for user behavior simulation fall short in several key aspects.
First, most datasets record only sparse, decontextualized--or even synthetic--user actions.
Some shopping datasets like Amazon-M2 or ECInstruct~\cite{jin2024amazon,jin2024shopping} record only the isolated actions (e.g., purchases or clicks) with limited observation context. Others~\cite{deng2024mind2web, chen2024gui, yao_webshop_2022} use synthetic or third-party annotated behaviorial data, which lacks the individual behavior pattern and the authenticity.
Additionally, few datasets provide step-level reasoning or persona information,
despite prior work has shown that rationale can improve LLM agent's performance in behavior and decision modeling~\cite{lu2025believabilityaccuratehumanbehavior,deepseek-aiDeepSeekR1IncentivizingReasoning2025}. Similarly, user persona strongly correlates with behavioral patterns~\cite{helmi2023characterizing}, making persona essential for durable personalization experience.

To address these limitations, we introduce \datasetname, a dataset of \textbf{O}bservation, \textbf{Pe}rsona, \textbf{R}ationale, and \textbf{A}ction collected from real human users during online shopping.
\datasetname provides rich, time-aligned logs of users' web browsing behavior, completed with self-reported rationales and detailed self-reported persona profiles. 
Unlike prior datasets, \datasetname captures not only \textbf{what} users do but also \textbf{why} they do it, enabling deeper insights into decision-making processes.
This paper focus on the online shopping domain as a beginning point due to its everyday prevalence, complex decision flow, and strong ties to personalization~\cite{
mican2020analysis, wei2016decision,
zhang2025shop,wang2025customer,zhang2025see}. 
Online e-commerce environments like Amazon requires a user's multi-step interactions involving comparisons, trade-offs, and goal-directed behaviors in one shopping session, all of which are prime testbeds for studying the capacity of LLMs to simulate real human actions.

To collect the \datasetname dataset, we developed \pluginname, a custom browser plugin that captures user interactions alongside corresponding web context and triggers rationale prompts at decision points, as shown in Figure~\ref{fig:firstpage}.
We also collect rich persona information through an online survey and an optional interview to include user profile information such as demographics, shopping styles, and personality traits.

The \datasetname-full contains 692 shopping sessions from 51 unique users, 28,904 <action, observation> pairs, and 604 human-annotated rationales. After post-processing, we also provide \datasetname-filtered, which includes 527 sessions, 5856 <observation, action> pairs and 207 rationales. \datasetname serves as the first benchmark dataset for evaluating LLM agents on \textbf{personalized} and \textbf{verifiable} user behavior simulation.
We benchmark four state-of-the-art LLMs (GPT-4.1~\cite{openai2025gpt41}, DeepSeek-R1~\cite{deepseek-aiDeepSeekR1IncentivizingReasoning2025}, Claude-3.7~\cite{anthropic2025claude37}, and Llama-3.3~\cite{meta2024llama33}) on \datasetname-test (a subset of the \datasetname-filtered) and analyze their ability to predict the next action and rationale of a specific user based on their persona and interaction history. 
These findings lay a foundation for future work on building LLM-powered digital twins capable of accurate and adaptive behavior modeling.

\begin{table*}
  \centering
  \begin{booktabs}{
  colspec={X[l]ccccccc},
  cells={m,font=\footnotesize},
  row{1}={font=\bfseries\footnotesize},
}
\toprule
Dataset        & Size & Task   & O  & Pe & R & A & Source \\
\midrule
Amazon Review  & 571M & Review Prediction                       & \cross       & \cross      & \cross    & Purchase           & User     \\
ECInstruct-SA  & 10k  & Sentiment Analysis                      & \cross      & \cross      & \cross   & Purchase                                      & User       \\
ECInstruct-REC & 10k  & Recommendation                          & \cross    & \cross      & \cross    & Purchase                                      & User        \\
Amazon-M2      & 3M   & Recommendation                          & \cross     & \cross      & \cross     & Click                                         & User      \\
Repeat Buyers  & 54M  & Buyer Prediction                        & \cross    & \checkmark      & \cross     & {Click, Cart  \\ Favor, Purchase}                   & User       \\
Taobao         & 100M & Recommendation                          & \cross      & \cross      & \cross     & {Click, Cart\\ Favor, Purchase }                  & User     \\
YOOCHOOSE      & 9M   & Purchase Prediction                     & \cross     & \cross      & \cross       & Click, Purchase                               & User    \\
Shopping MMLU  & 3973 & Recommendation                          & \cross     & \cross      & \cross      & {Query, Click\\ Purchase}             & User     \\
\midrule
Mind2Web       & 2350 & Web Navigation                          & \checkmark    & \cross      & \cross     &{ Click, Hover, \\ Type, Select }                    & {Annotator,\\ GPT}     \\
GUI-WORLD      & 12k  & {GUI Understanding \\ Instruction Follow}    & \checkmark   & \cross      & \cross     & {Click, Paste,\\ Search, Type}                     & {Annotator,\\ Video}    \\
WebArena       & 812  & {Web Navigation \\Instuction Follow}        & \checkmark   & \cross      & \cross     & {Click, Hover, \\Type, Tab Switching, \\Navigation }& {Annotator,\\ GPT}     \\
WebShop & 1600 & Web Navigation & \checkmark  & \cross & \cross & Input, Click & Annotator\\
\midrule
 {\datasetname-full \\ \datasetname-filtered} & {692\\527}  & {All Above and \\ User Behavior Simulation} & \checkmark   & \checkmark      & \checkmark     & {Basic Action,\\ Semantic Action}                 & User      \\
\bottomrule
\end{booktabs}

  \caption{Properties of existing datasets compared to \datasetname. ``O'': Environment Observation. ``Pe'': User Persona. ``R'': Rationale behind action. ``A'': Action Space. ``Source'': Action Source. }
  \label{tab:dataset}
\end{table*}

\section{Related Works}
\subsection{LLMs for Human Behavior Simulation}
Large Language Model agents can handle complex tasks~\cite{wang2024sotopia,yao2023react,shinn2023reflexion,wu2023autogen,zhang2024revisiting,jia2024soul,chen2025unlearning,wang2026trajectory2task}, and researchers are using them as human proxies across domains, from social-science simulations~\cite{park2023generative,wang2024userbehaviorsimulationlarge,10.1145/3708359.3712149} and recommender-system evaluation~\cite{wang2023recmind} to UX testing~\cite{lu2025uxagentllmagentbasedusability} and health-counsellor training~\cite{louie2025llmsimulatedpracticefeedbackupskill}. They even reproduce classic results in experimental psychology and economics, such as Milgram Shock Experiment~\cite{10.5555/3618408.3618425}.

In parallel, there have been works looking at generating synthesized personas based on text data, such as PersonaHub~\cite{ge2024scaling} and the approach by ~\citet{shi2025you}, demonstrating promise in downstream modeling.
Moreover, several approaches have integrated persona information to enrich behavioral simulation~\cite{shao2023character,chuang2024beyond,shi2025you}. For example,~\citet{parkGenerativeAgentSimulations2024} introduces an persona-grounded framework using qualitative interviews, enabling agents to accurately simulate individual preferences, attitudes, and behaviors, showing that incorporating personas improves the realism of simulated behavior.

Moreover, there is a growing trend for using LLM agents in online scenarios to produce human-like behaviors, such as Claude Computer Use~\cite{claude35}, AutoGLM~\cite{liu2024autoglm}, Coco-Agent~\cite{ma2024coco}, Mobile-Agent-E~\cite{wang2025mobile}, and OpenAI Operator ~\cite{openai2025operator}, enabling interaction in more complex human-computer settings. 
Yet, existing research on LLM web agents predominantly focuses on optimizing and evaluating agents for task completion~\cite{gurRealWorldWebAgentPlanning2023, zhouWebArenaRealisticWeb2024, he-etal-2024-webvoyager}, training agents to learn the fewest, most direct steps--whereas real users behaviors normally contain richer and nonlinear paths. 
There remains a lack of work exploring accurate simulation of human behavior, particularly the modeling of personalized user behaviors.
Motivated by this limitation, we propose \datasetname, which aims to better facilitate research on simulating realistic and personalized human behaviors in online scenarios.

\begin{figure*}[t]
    \centering
    \includegraphics[width=\textwidth]{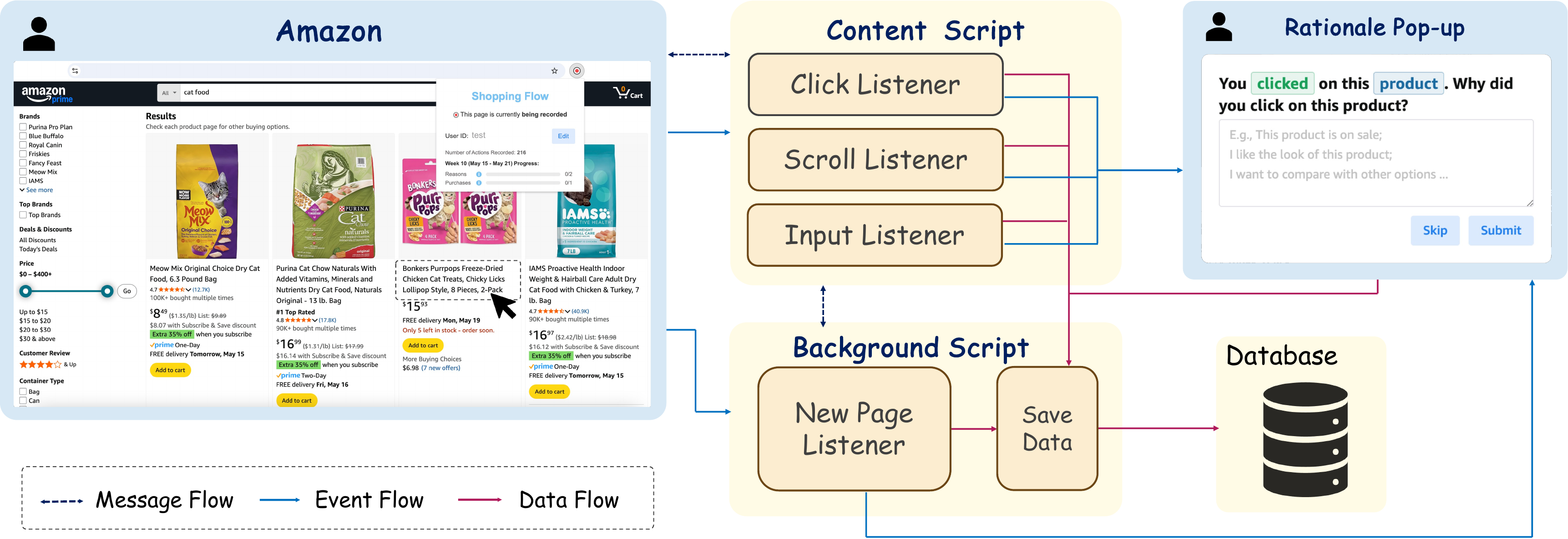}
    \caption{Pipeline of our Chrome Plugin, \pluginname. \texttt{Content Script} detects click, scroll and input actions. \texttt{Background Script} detects page-related actions and handles data uploading. Rationale pop-up is triggered at a certain probability when certain action types are detected.}
    \label{fig:pipeline}
\end{figure*}
\subsection{User Online Behavior Datasets}

Existing datasets containing user online behaviors can be broadly categorized into two sets, shown in Table~\ref{tab:dataset}.
The first category consists of recommendation-oriented datasets which mostly records user interaction with product items (e.g., click, purchase), lacking of essential context.
A commonly used example is the \texttt{Amazon Review 2023} dataset \cite{hou2024bridging}, which contains over 571 million user review-writing behaviors and links them with product information. 
Follow-up studies, such as \texttt{ECInstruct} \citep{peng2024ecellm}, have utilized the Amazon Review to work on tasks like user review sentiment analysis. 
Others, such as \texttt{Amazon-M2} \cite{jin2024amazon}, \texttt{YOUCHOOSE} \cite{ben2015recsys}, \texttt{Repeat Buyers}~\cite{liu2016repeat}, and \texttt{Taobao} \cite{zhu2018learning}, capture richer shopping behaviors such as clicks, add-to-cart, and purchases. The \texttt{Repeat Buyers} dataset further offers basic user persona like gender and age.
However, these datasets lack fine-grained user behaviors and contextual information needed to explain how actions and decisions are made at each stage of the user shopping journey, limiting their utility in user behavior simulation.

A separate line of datasets focuses on task completion, with environment observations for web agents tasks, such as \texttt{Mind2Web}~\cite{deng2024mind2web}, \texttt{GUI-WORLD}~\cite{chen2024gui}, \texttt{WebArena}~\cite{zhou2024webarenarealisticwebenvironment}, and \texttt{WebShop}~\cite{yao_webshop_2022}. These datasets contain user interaction traces on website or mobile devices (e.g., clicks, typing) and HTML or screenshots. While useful for studies like instruction following, these datasets are built from annotator-generated or synthetic interaction logs. As a result, they lack both behavioral authenticity and personalization, limiting their realism.

There are other datasets and benchmarks which focus on evaluating LLM agents' reasoning ability using question-answering format, for example~\citet{mialon2023gaiabenchmarkgeneralai}. While such tasks are valuable for evaluating models, these datasets are not suitable for simulating human behavior.

In summary, existing datasets capture certain aspects of user behaviors but lack the comprehensive data required for simulating real user behavior.

\section{\datasetname Dataset}

Our work presents a novel dataset \datasetname to advance NLP research in realistic user behavior simulation, particularly within the context of online shopping scenarios.

\subsection{\datasetname Data Collection}
\subsubsection{Participant Recruitment}
A total of $84$ participants were recruited through snowball sampling~\cite{goodman1961snowball}.
During the pre-screening survey, candidate participants were required to self-identify if they were frequent Amazon customers and if they planned to make at least one purchase on Amazon in the next several weeks.
All participants were required to install the \pluginname Chrome plugin (details in Section~\ref{sec:shopping-behavior-collection}), shop normally on the Amazon website for a four-week period, and participate in an online survey and an optional interview.
Details of recruitment are in Appendix~\ref{dataCollectionDetail}.

\subsubsection{Persona Information Collection}
We collected detailed user persona information through a structured online survey and an optional semi-structured interview.
All survey questions are designed based on established work (design details in Appendix~\ref{survey-detail}) to solicit consumer characteristics that have been shown to correlate with shopping behaviors. 
The survey consists of three main sections: \textbf{demographic information}, \textbf{shopping preferences}, and \textbf{personality traits}. 
Demographic information includes age, gender, education, occupation, income, residence, and self-description. 
Shopping preferences includes online shopping frequency, membership status, shopping habits, seasonality, advertising trust, review engagement, delivery influence, and an adapted eight-item Consumer Styles Inventory (CSI)~\cite{nayeem2022revisiting}. 
Personality traits are measured using the Big-Five Inventory~\cite{goldberg1992development} and a self-reported MBTI personality~\cite{myers1998}. 

Following prior work on persona information collection~\cite{parkGenerativeAgentSimulations2024}, we invited participants to attend a 20-minute optional semi-structured interview about their \textbf{demographic information, personal background, and online shopping preferences}, which aimed to better contextualize the personalized decision-making processes behind users' online shopping actions.
The interview design description and the protocol used are provided in Appendix \ref{Interview-protocol}.

\begin{figure*}[t]
    \centering
    \includegraphics[width=\textwidth]{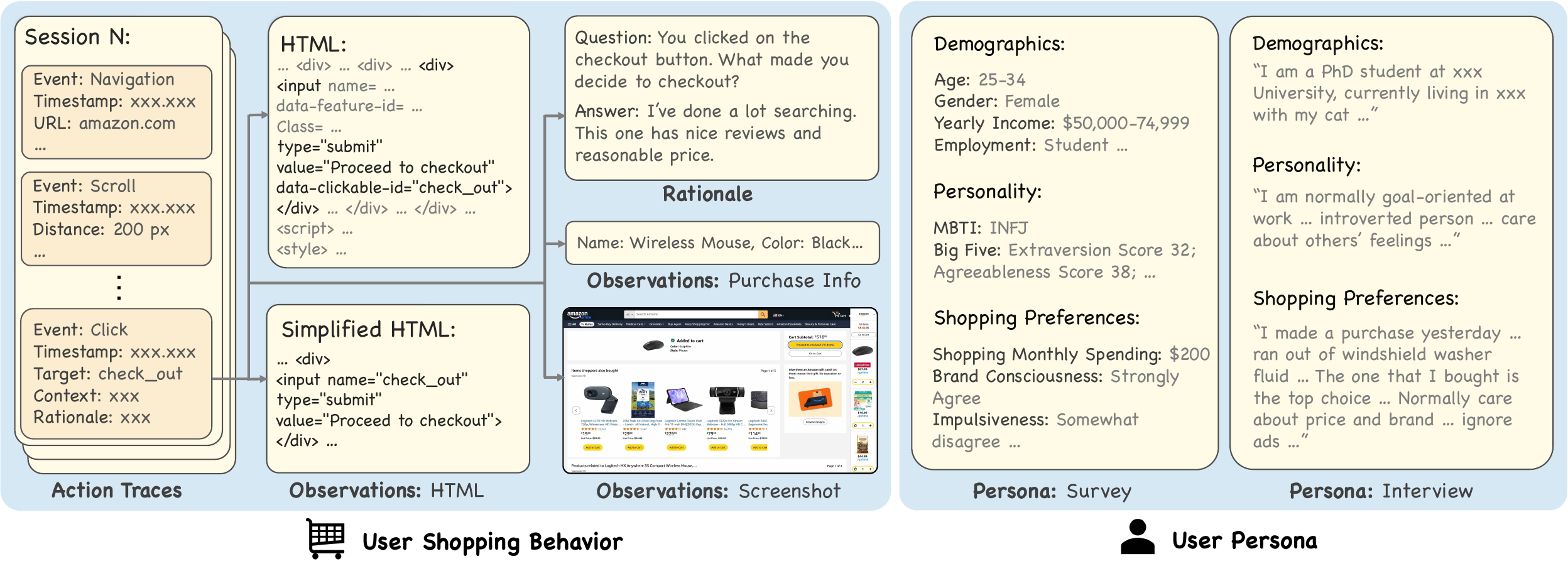}
    \caption{\datasetname Dataset Overview. The dataset comprises four major components: \textbf{action traces}, \textbf{web observations}, \textbf{rationales}, and \textbf{user personas}. 
 Each shopping session is a sequence of timestamped actions. Each action is paired with a corresponding web observation, which includes: the full HTML of the interacted webpage, a simplified HTML with key elements, a screenshot, and product information for purchases (if applicable). 
The rationale is a natural language explanation of why the user performed the action. 
The persona contains detailed user profiles collected from surveys and interviews, covering demographics, personality traits, and shopping preferences.}
\label{fig:data-overview}
\end{figure*}
\subsubsection{Shopping Behavior Collection}
\label{sec:shopping-behavior-collection}
To support the construction of the dataset, we designed \pluginname, a Chrome extension that automatically captures user behaviors and contextual web observations during Amazon shopping sessions (shown in Figure~\ref{fig:pipeline}).

The plugin includes two main scripts: a \texttt{Content Script} that runs within the Amazon page to log user interactions (including inputs, clicks, and scrolls) with timestamps, target elements, and HTML using custom parsing rules (Appendix~\ref{recipe-design}); and a \texttt{Background Script} that tracks page-level events like reloads and navigation. All data was securely uploaded to Amazon S3.
To capture rationale, the plugin randomly triggers pop-ups (8\% chance) asking users to explain their actions (question design in Appendix~\ref{reasoning-design}).

\subsubsection{Post-Processing}
To ensure data quality and to \textbf{protect user privacy, }we applied a multi-step post-processing procedure.
We configured the plugin to not record any personally identifiable information (PII), such as the user login page, account profile page, or the checkout details. In addition, we designed a rule-based automated detection and pattern matching script to mask any PII unavoidably contained in a page (e.g., username in navigation bar), including usernames, zip codes, addresses, specific workplaces, and payment details, before any human touch.
Lastly, we manually checked the data to ensure there is no PII in the dataset.

The actual purchase information is not collected since it is in the checkout detail page. Instead, we infer a ``purchase'' action via click actions on ``proceed to checkout,'' ``buy now,'' and ``set subscription'' buttons. 
We associate the inferred purchase action with the corresponding product information during data post-processing.

In addition, the raw user data is a stream of continuous user action sequences that do not separate different shopping sessions.
Thus, our team segmented user actions using a two-step strategy based on temporal intervals and purchase signals. 
Sessions were first split using a time threshold and then further segmented at purchase intention events (i.e., clicking on ``proceed to checkout / buy now / set subscription / add to cart'' button). 
Detailed explanation about the selection of the threshold can be found in Appendix~\ref{intervalDistribution}. 
Finally, sessions with fewer than five actions were discarded to remove trivial or non-informative behaviors as a prior work reported meaningful sessions typically contain at least six to seven interactions~\cite{wang2025agenta}.

To reduce noise and improve the quality of behavioral data, some actions that occurred on uncommon or rarely visited pages or do not reflect meaningful intent are removed. 
Moreover, we filtered out clicks on non-interactive areas such as the background, and further filtered actions involving Amazon Rufus.

To support behavior modeling and evaluation under a more tractable setting, we follow prior work~\cite{lu2025believabilityaccuratehumanbehavior} and similarly define a simplified action $\mathcal{A}$ space consisting of key interaction types. Specifically, we retain three high-level actions that are both semantically meaningful and commonly observed: input, click, and terminate. Within the ``click'' category, we further differentiate between several subtypes, shown in Table ~\ref{tab:click_statistics}. This abstraction reduces complexity while preserving the structure of user behaviors, enabling more stable behavior simulation.

\subsection{Dataset Details}
An overview of the dataset is presented in Figure~\ref{fig:data-overview}.
Of all the users, 51 contributed at least one shopping session. In total, the \datasetname-full contains 692 sessions, 28,904 \textless action, observation\textgreater\ pairs, and 604 rationale annotations. The \datasetname-filtered contains 527 sessions, 5,856 \textless action, observation\textgreater\ pairs, and 207 rationales.
Table~\ref{tab:data_stats} presents the statistics. Table~\ref{tab:click_statistics} shows the click type distribution.

\paragraph{User Persona}
For participant $i$, the persona is represented by $P_i$, comprising two components: a structured survey and interview. Both focusing on their demographics, personality or personal background, and shopping preferences.

\paragraph{User Action Traces}
Each shopping session $j$ includes users' web interactions on the shopping website. The action trace is represented by $A_j = \{a_1, ..., a_{T}\}$, where $a_t \in \mathcal{A}$ represents the user’s action at step $t$ in the action space $\mathcal{A}$. The Action space $\mathcal{A}$ includes \texttt{click, scroll, input, navigate, tab activate}.
Each action is assigned a unique identifier (UUID) and is timestamped to preserve the exact temporal order. 

In particular, for each click action,  the corresponding CSS selector of the element is provided to uniquely identify the clicked target. 
Additionally, semantic identifiers are assigned to click actions to indicate their functional context, such as interactions with products in search results (e.g., \texttt{semantic\_id: “search\_result.product\_name”}) or interactions with other commonly used page elements. This semantic id enables downstream models to have a clearer recognition and observation of user behaviors when utilizing the dataset. 
Similarly, scrolling actions are presented with their start and end positions, supporting analyses of how users explore page content. 

\paragraph{Rationale}
Alongside the user actions, rationales are provided for some actions, $R_j = \{r_1, \dots, r_{T}$, where $r_k$ is a nullable string describes the user's rationale for specific action $a_k$, explicitly capturing the underlying motivations or thought processes of users. These insights enable the downstream models to have a thorough understanding of why users make particular choices, offering a deeper view of user behavior and the reasoning process.

\paragraph{Web Observation}
In addition to action traces and rationales, the observation of the web context at each action step in session $j$ is captured, represented by $O_j = \{o_0, ..., o_{T}\}$. Each $o_t$ includes the \textbf{HTML} content and a \textbf{screenshot}\footnote{The experiments presented in this paper did not involve the use of screenshot data.}. The HTML content also contains annotations indicating the viewability of each clickable element, as well as relevant page metadata (such as product name, product price, etc.) at time $t$. 
To reduce noise and storage while keeping the HTML structure, we provides the a simplified version of the HTML content containing key elements for frequently encountered pages, such as search results, product detail pages, and shopping carts. 

Additionally, if a purchase is made during a session, the final purchase information, including the price, product title, product options, and Amazon Standard Identification Number (ASIN), are recorded to facilitate potential downstream tasks such as recommendation.

\begin{table}[t]
\centering
\begin{booktabs}{
  colspec={X[l]ll},
  cells={font=\small},
  row{1}={font=\bfseries},
}
\toprule
Action Type & \# of Full & \# Filtered \\
\midrule
Scroll        & 19,217 (66.5\%) & --  \\
Click         & 5,253  (18.1\%) & 5,051 (86.3\%)  \\
Tab Activate  & 1,945  (6.7\%)  & --   \\
Navigate      & 1,901  (6.6\%)  & --    \\
Text Input         & 606    (2.1\%)  & 597  (10.2\%)         \\
Terminate & -- & 208(3.6\%)\\
\midrule
Total         & 28,904          & 5,856            \\
\bottomrule
\end{booktabs}
\caption{Action type distribution.}
\label{tab:data_stats}
\end{table}

\begin{table}[t]
\centering
\begin{booktabs}{
  colspec={lcc},
  cells={font=\small},
  row{1}={font=\bfseries},
}
\toprule
\textbf{Click Type} & \textbf{Count} & \textbf{Percentage} \\
\hline
review             & 1052 & 20.8\% \\
search             & 763  & 15.1\% \\
product\_option     & 700  & 13.9\% \\
product\_link       & 537  & 10.6\% \\
other              & 449  & 8.9\%  \\
purchase           & 321  & 6.4\%  \\
nav\_bar            & 283  & 5.6\%  \\
page\_related       & 198  & 3.9\%  \\
quantity           & 191  & 3.8\%  \\
suggested\_term     & 182  & 3.6\%  \\
cart\_side\_bar      & 145  & 2.9\%  \\
cart\_page\_select   & 139  & 2.8\%  \\
filter             & 91   & 1.8\%  \\
\bottomrule
\end{booktabs}
\caption{Click type distribution in \datasetname-filtered. Detailed descriptions in Appendix~\ref{app:click_type_explanation}}
\label{tab:click_statistics}
\end{table}

\begin{table}[t]
\centering
\begin{booktabs}{
row{1}={font=\bfseries},
}
\toprule
Metric & Value\\
\midrule
\# of Session &  527\\
Avg. \# of Action & 11.11\\
Avg. \# of Input &1.13 \\
Avg. \# of Click & 9.58 \\
Avg. \# of Terminate &0.39\\
\bottomrule
\end{booktabs}
\caption{\datasetname-filtered dataset statistics per-session}
\label{tab:processed_data_1}
\end{table}

\section{Tasks and Experiments}

\subsection{Tasks}
We show how \datasetname can be leveraged to evaluate LLM's ability to simulate consumer behavior in online shopping. 

\paragraph{Next Action Prediction}
The next action prediction task aims to model the user's next action based on previous behaviors following the definition of previous work~\citep{deng2024mind2web,lu2025believabilityaccuratehumanbehavior}. Given a history of actions $\{a_1,\cdots,a_{t-1}\}$ in shopping session $j$, corresponding web contexts $\{o_1, ..., o_{t}\}$, rationale $\{r_1, ...,r_t\}$\footnote{Note: Rationale annotations are sparse, whereas actions and observations are fully recorded at each time step.}, and the consumer profile $P_i$, the model is tasked to predict the immediate next action $a_t$, learning a function of the form:
    \[
    a_t = F_{action}(a_{1...t-1}, r_{1...t-1} ,o_{1...t}, P_i) 
    \]


\subsection{Experiments}

\subsubsection{Experiment Setup}
From the \datasetname-filtered, we furthur construct the test set \textbf{\datasetname-test} by randomly sampling 15 out of 51 users and randomly sampling 90 sessions from these users, resulting in 992 actions. We evaluate four state-of-the-art LLMs, including two open-source models, Llama-3.3-70B-Instruct and DeepSeek-R1, and two proprietary models, GPT-4.1 and Claude-3.7-Sonnet.

All models are evaluated in a zero-shot, prompt-based setting without fine-tuning. Prompt templates are provided in Appendix~\ref{prompt}.

To investigate how different input factors affect model behavior, we conducted a series of ablation studies. Specifically, we excluded persona information (\texttt{w/o persona}) and additionally removed the history rationale from the input (\texttt{w/o rationale}).

\begin{table*}[t]
\centering
\begin{booktabs}{
  colspec={lc|ccc},
  cells={font=},
  row{1}={font=\bfseries},
}
\toprule
\textbf{Model} &
\makecell{Next Action Gen.\\(Accuracy)} &
\makecell{Action Type\\(Macro F1)} & 
\makecell{Click Type\\(Weighted F1)} &
\makecell{Session Outcome\\(Weighted F1)} \\
\midrule
GPT-4.1 & 21.51  &\textbf{48.78} & \textbf{44.47} & 47.54 \\
\hspace{0.4em}w/o persona & \textbf{22.06}  & 45.55 & 43.45 & \textbf{58.47} \\
\hspace{0.8em}w/o rationale & 21.28 & 34.93 & 42.63 & 51.17 \\
\midrule
DeepSeek-R1 & 14.75  & 27.37 & 35.12  & 46.36 \\
\hspace{0.4em}w/o persona & 15.52  & 27.43 & 33.86  & 48.86 \\
\hspace{0.8em}w/o rationale & 15.74  & 27.16 & 32.65  & 47.92 \\
\midrule
Claude-3.7 & 10.75  & 31.58 & 27.27  & 43.52 \\
\hspace{0.4em}w/o persona & 10.75  & 25.33 & 22.76 & 43.10 \\
\hspace{0.8em}w/o rationale & 10.08  & 26.06 & 20.29  & 43.10 \\
\midrule
Llama-3.3 & 8.31  & 24.29 & 19.99  & 36.64 \\
\hspace{0.4em}w/o persona & 8.31  & 23.69 & 18.59  & 33.21 \\
\hspace{0.8em}w/o rationale & 8.76  & 23.60 & 19.22 & 34.19 \\
\bottomrule
\end{booktabs}
\caption{
Evaluation of actions in next action prediction task. We report four metrics here to assess model performance (Full results can be found in Appendix~\ref{app:action_prediction_full_result}): Next Action Generation Accuracy measures the exact-match accuracy of the predicted next action; Action Type Macro F1 evaluates the model’s ability to predict the correct high-level action category (e.g., input, click, terminate); Click Type Weighted F1 captures the performance of predicting the specific type of click actions; Session Outcome Weighted F1 reflects how well the model can predict the final outcome of a session, where each session ends either in a purchase or a terminate action.
``Claude-3.7'': Claude-3.7-Sonnet, ``Llama-3.3'': Llama-3.3-70B-Instruct. All metrics are reported as percentages (\%). Instance size $n=902$. 
}
\label{tab:action_predict}
\end{table*}

\subsubsection{Evaluation}

To assess the accuracy of generated user actions, we apply an exact match criterion: a prediction is correct only when all required components align with the ground truth.
Specifically, for click actions, the clicked target name must match.
For input actions, this includes identifying the input field and generating exact input text.

In addition to exact match accuracy, we evaluate the model’s ability to classify action types. We report weighted F1 for high-level action types (\texttt{click}, \texttt{input}, \texttt{terminate}). Given the highly imbalanced nature of user behavior distributions, we also report macro F1 to highlight performance across all classes regardless of frequency. 

To further examine fine-grained prediction capabilities, we evaluate the weighted F1 score for \texttt{click} subtypes. This captures whether the model not only predicts that a user will click but also understands the specific type of click (e.g., \texttt{review}, \texttt{product\_link}, \texttt{purchase}).

Finally, given the goal-oriented nature of online shopping, we assess the model's ability to predict session outcomes. Each session ends in either a \texttt{click on purchase-related button} or a \texttt{terminate} action. We evaluate the model’s accuracy and F1 score on these terminal actions to understand whether it can correctly capture users’ long-term goals and decision-making processes.

\subsubsection{Results and Analysis}

\subsection{Main Results}
The results for next action prediction are presented in Table~\ref{tab:action_predict_filtered}. 
Among all models, GPT-4.1 achieves the strongest overall performance. It obtains the highest action generation accuracy of 22.06\%, a macro F1 score of 48.78\% on action type prediction, and a weighted F1 score of 44.47\% on click type prediction. For session outcome prediction, GPT-4.1 also shows solid performance with 58.47\% F1. This strong performance may be attributed to its large context window, which facilitates better processing of complex interaction histories.
DeepSeek-R1 performs moderately well across tasks. It achieves a maximum action generation accuracy of 15.74\% and obtains solid outcome prediction, possibly due to its strong reasoning abilities. However, the performance may be limited by the model's 128k context length.
Claude-3.7 shows modest performance. Its action generation accuracy is around 10.75\%, and action type prediction is generally weaker than GPT-4.1 and DeepSeek. Nonetheless, it achieves relatively good outcome prediction, suggesting some robustness in capturing high-level user intent.
LLaMA-3.3 underperforms across all metrics, with action generation accuracy of only 8.76\%. Its lower performance may be due to the smaller model size (70B) and shorter context length (128k).

\begin{table*}[t]
\centering
\begin{tabular}{lcccc}
\toprule
\textbf{Error Type} & \textbf{GPT} & \textbf{R1} & \textbf{Claude} & \textbf{LLaMA} \\
\midrule
Didn't Terminate & 35 (3.9\%) & 39 (4.3\%) & 40 (4.4\%) & 40 (4.4\%) \\
Didn't Click & 49 (5.4\%) & 21 (2.3\%) & 33 (3.7\%) & 27 (3.0\%) \\
Didn't Input & 50 (5.5\%) & 70 (7.8\%) & 55 (6.1\%) & 74 (8.2\%) \\
Input Wrong Field & 0 (0.0\%) & 0 (0.0\%) & 1 (0.1\%) & 0 (0.0\%) \\
Input Wrong Text & 26 (2.9\%) & 6 (0.7\%) & 19 (2.1\%) & 2 (0.2\%) \\
Click Wrong Button & 548 (60.8\%) & 633 (70.2\%) & 657 (72.8\%) & 684 (75.8\%) \\
\bottomrule
\end{tabular}
\caption{Error type breakdown across models with count and percentage.}
\label{tab:error_breakdown}
\end{table*}

\begin{table*}[t]
\centering
\begin{tabular}{lcccccc}
\toprule
\textbf{Action Type} & \textbf{Ground-Truth} & \textbf{GPT} & \textbf{R1} & \textbf{Claude} & \textbf{LLaMA} \\
\midrule
Click      & 786 (87.14\%)  & 819 (90.80\%) & 865 (95.90\%) & 843 (93.46\%) & 862 (95.57\%) \\
Input      & 76 (8.43\%)    & 54 (5.99\%)   & 21 (2.33\%)   & 48 (5.32\%)   & 3 (0.33\%) \\
Terminate  & 40 (4.43\%)    & 29 (3.22\%)   & 5 (0.55\%)    & 0 (0.00\%)    & 0 (0.00\%) \\
Other      & 0 (0.00\%)     & 0 (0.00\%)    & 11 (1.22\%)   & 11 (1.22\%)   & 37 (4.10\%) \\
\bottomrule
\end{tabular}
\caption{Distribution of predicted and ground truth action types with count and percentage.}
\label{tab:action_distribution}
\end{table*}

The role of persona information varies across models. While adding persona information does not consistently improve exact action generation accuracy, it generally enhances the model’s performance on action type and click type prediction across. This suggests that persona information provides useful priors about user preferences and behavior patterns, helping the model better classify action semantics. However, its limited effect on action generation accuracy implies that simply including persona in the prompt may introduce noise rather than help. Current models have limited ability to deeply integrate persona into step-level decision-making. This highlights potential room for improvement in personalized user modeling.

Additionally, removing historical rationales consistently leads to performance degradation across most models and metrics, particularly in outcome prediction. This confirms that rationale information serves as valuable intermediate supervision, guiding the model to align its decisions with plausible user intent. We also note several outliers in the results. For instance, LLaMA-3.3 does not consistently benefit from rationale inputs. This may be due to its smaller model size and limited capacity to leverage additional contextual signals effectively.

\subsection{Error Analysis}

As shown in Table~\ref{tab:error_breakdown}, the majority of model failures are attributed to incorrect button click predictions. In addition, models frequently struggle to accurately generate input or termination actions. Even in cases where the model successfully identifies an input action, it often fails to reproduce the correct search query. 

Table~\ref{tab:action_distribution} further highlights the discrepancy between the predicted and ground-truth distributions of action types. Notably, the “terminate” action, despite its presence in the ground-truth data, is rarely predicted by most models (except for GPT-4.1). This mismatch suggests a potential bias in current LLMs to be optimized for completing the shopping task (i.e., the purchase), rather than simulating realistic user behavior, which often includes early session termination.

\section{Conclusion}

This paper introduces \datasetname, a comprehensive online shopping behavior dataset specifically designed to advance the development and evaluation of LLM-based agents for simulating user behavior. 
By capturing full shopping trajectories, including action traces, web observations, user personas, and explicit rationales, the dataset provides a verifiable, personalized resource for user behavior modeling. 
We define a suite of evaluation tasks and conduct a comprehensive analysis across four state-of-the-art LLMs.
Our results highlight both the promise and current limitations of LLM agents in simulating realistic user behavior. 
While certain models demonstrate plausible rationale and action prediction under simple setups, there remains substantial room for improvement, especially in handling complex decision flows and deeper personalization.

\section*{Limitations}
This study evaluates LLM agents under a simplified setup, following prior work~\cite{lu2025believabilityaccuratehumanbehavior}, adopting a reduced action space and a coarse-grained session segmentation strategy. In particular, actions such as scrolling and page navigation are omitted. This simplification serves to manage the complexity of the task: while text-based LLMs are robust in processing structured language data, they might struggle to accurately model continuous UI-based actions like scrolling. We hope that future models capable of richer, multimodal reasoning may better handle these complexities.

Second, although we collect screenshots alongside HTML data for every user interaction, the experiments in this paper do not incorporate visual signals. This is primarily because existing LLM agents are not yet robustly equipped to interpret raw visual UI elements from screenshots in conjunction with structured web content. We envision that future work can leverage this visual information to interpret user decisions.

Looking forward, the \datasetname dataset enables a wide range of future research directions beyond those explored in this paper (e.g. personalized recommendation). Furthermore, simulation based on \datasetname offers a scalable framework for generating synthetic interaction data, which could be valuable for applications like website design and adaptive user interaction. We hope this dataset serves as a foundation for more realistic, personalized, and interpretable user behavior modeling.

\bibliography{custom,yuxuan}

\clearpage

\appendix

\section{Data Collection Implementation Details}
\label{dataCollectionDetail}
We recruited 84 participants via snowball sampling. The participants are pre-screened to ensure that they meet the following criteria: at least 18+ years old, based in the U.S., English speaker and have used (or plan to use) Amazon website to make a purchase in the past (future) couple of weeks. 

The participant incentive structure consists with: a) \$5  for completing the online survey (10 mins) and b) \$10 for participating in an optional 20-mins interview to discuss about their personal background. c) \$5 per one qualified week for participants who have complete one or more purchase sessions, with at least two rationale recorded by the \pluginname plugin. In addition, we provide \$10 as a bonus to participants who can successfully complete four or more qualified weeks of data collection in a row. 
All incentives were delivered as Amazon digital gift cards to their emails. 

During the data collection process, no personally identifiable information was retained. To ensure data privacy, we don't collect data on sensitive pages like checkout or account page. Meanwhile, we implemented a script to anonymize sensitive details in recorded context data, such as zip codes, names, and addresses. Any screenshots that unintentionally contained identifiable information were reviewed and removed prior to dataset release. Furthermore, a research assistant conducted a post-screening review to confirm the exclusion of identifiable information.

This study was conducted in compliance with the Institutional Review Board (IRB) guidelines at Northeastern University.

\section{Survey Design}
\label{survey-detail}
The survey consists of three main sections: \textbf{demographic information}, \textbf{shopping preferences}, and \textbf{personality traits}.

Demographic information significantly influences consumer behavior~\cite{hou2021mobile}. The survey collects age, gender, education level, occupation, family income, location of residence, and a two-sentence self-description.

Shopping preferences section asks for participants' online shopping frequency and whether they have a paid membership or not. 
In addition, we include 12 questions inspired by previous literature, all with a 5-point Likert scale answer \cite{likert1932technique}.
These questions include four shopping habits items, seasonality \cite{mastercard_holiday_2023},  
tendency of believing in advertisements~\cite{nizam2018interactive}, habits of reading product reviews~\cite{shao2014impact}, and the influence of delivery~\cite{bauboniene2015commerce}  
and an 8-items consumer styles inventory (CSI) adapted to the online shopping context ~\cite{sprotles1986methodology,helmi2023characterizing,prakash2018application,nayeem2022revisiting}, 
including: brand loyalty, price conciousness, perfectionism and high-quality conciousness, impulsiveness, confusion by overchoice, brand consciousness, recreational consciousness, and novelty and fashion conciousness.

Personality traits section 
utilizes the Big-Five Personality Inventory with five core dimensions: Openness to experience, conscience, extrovertism, agreeableness, and neuroticism~\cite{goldberg1992development}. A self-reported MBTI is also included~\cite{myers1998}.
\subsection{Survey Questions}
\subsubsection{Demographic Information}
\noindent Q1: Gender.

[Male; Female; non-binary]
\\
\noindent Q2: Age.

[Under 18; 18-24; 25-34; 35-44; 45-54; 55-64; 65+]
\\
\noindent Q3: Which city do you live?

\noindent Q4: What is your highest level of education ?

[High school diploma or lower; Bachelors' degree or current college student; Graduate degree or current grad student (MA, MS, MBA, etc.); Doctoral degree or current doctoral student (PhD, JD, MD, DDS etc.); Prefer not to say]

\noindent Q5: Do you live alone or live together with others, if so who are they? (Optional)

\noindent Q6: What was your household before-tax income during the past 12 months? If you are a student, what's your allowance or stipend?

[Less than \$25,000; \$25,000-\$49,999; \$50,000-\$74,999; \$75,000-\$99,999; \$100,000-\$149,999; \$150,000 or more; Prefer not to say]

\noindent Q7: What best describes your employment status over the last three months?

\noindent Q8: What best describes your employment status over the last three months?  

[Full-time employee; Part-time employee; Self-employed; Unemployed and looking for work; Student; Retired; Other: 

\noindent Q9: Use two sentences to describe yourself.  Example 1:``I am a machine learning researcher at a startup focusing on autonomous driving. My daily work include developing and optimizing deep learning models for perception, sensor fusion, and decision-making in self-driving vehicles.''  Example 2: ``I am a PhD student in computer science, specializing in AI for healthcare. I frequently conduct experiments, analyze medical datasets, and collaborate with doctors to ensure our models are clinically interpretable. I also attend academic seminars, present my findings at conferences, and participate in lab meetings to refine research directions. '' 
\\

\subsubsection{Shopping Preferences}

\noindent Q10: How often do you shop online?  

[More than three times a week; Once to twice a week; Once every couple of weeks; Less than once a month]  
\\
\noindent Q11: How much money (in US dollars) do you spend on online shopping per month? (Not including food or delivery services)  
\\
\noindent Q12: Do you have a paid Amazon Prime membership?  

[Yes; No]  
\\

From Q13 to Q24, all items use the same response scale: [Strongly disagree, Somewhat disagree, Neither agree nor disagree, Somewhat agree, Strongly agree]\\
\noindent Q13: I tend to shop more during holidays (e.g.\ Black Friday, holiday sales).\\
\noindent Q14: Online ads attract my attention and are a good source of information.\\
\noindent Q15: I usually do a lot of research (e.g.\ reading online reviews) before making a purchase.\\
\noindent Q16: I prioritize delivery speed and delivery fee of the product.\\
\noindent Q17: Getting high-quality online products is very important for me.\\
\noindent Q18: The more expensive online product brands are usually my choice.\\
\noindent Q19: The more I learn about online products, the harder it seems to choose the best.\\
\noindent Q20: I shop quickly for online products, buying the first product or brand I find that seems good enough.\\
\noindent Q21: Once I find a brand I like, I stick with it.\\
\noindent Q22: I would buy a new or different brand of product just to see what it is like.\\
\noindent Q23: I enjoy shopping for online products just for the fun of it.\\
\noindent Q24: I look carefully to find the best value for money when shopping online.\\

\subsubsection{Personality Traits}
\noindent Big Five Test: read the statements carefully and indicate to what extent you agree of disagree. From Q25 to Q74, all items use the same response scale:
[Very Inaccurate, Moderately Inaccurate, Neither Accurate Nor Inaccurate, Moderately Accurate, Very Accurate]

\noindent Q25: Am the life of the party.\\
\noindent Q26: Feel little concern for others.\\
\noindent Q27: Am always prepared.\\
\noindent Q28: Get stressed out easily.\\
\noindent Q29: Have a rich vocabulary.\\
\noindent Q30: Don’t talk a lot.\\
\noindent Q31: Am interested in people.\\
\noindent Q32: Leave my belongings around.\\
\noindent Q33: Am relaxed most of the time.\\
\noindent Q34: Have difficulty understanding abstract ideas.\\
\noindent Q35: Feel comfortable around people.\\
\noindent Q36: Insult people.\\
\noindent Q37: Pay attention to details.\\
\noindent Q38: Worry about things.\\
\noindent Q39: Have a vivid imagination.\\
\noindent Q40: Keep in the background.\\
\noindent Q41: Sympathize with others’ feelings.\\
\noindent Q42: Make a mess of things.\\
\noindent Q43: Seldom feel blue.\\
\noindent Q44: Am not interested in abstract ideas.\\
\noindent Q45: Start conversations.\\
\noindent Q46: Am not interested in other people’s problems.\\
\noindent Q47: Get chores done right away.\\
\noindent Q48: Am easily disturbed.\\
\noindent Q49: Have excellent ideas.\\
\noindent Q50: Have little to say.\\
\noindent Q51: Have a soft heart.\\
\noindent Q52: Often forget to put things back in their proper place.\\
\noindent Q53: Get upset easily.\\
\noindent Q54: Do not have a good imagination.\\
\noindent Q55: Talk to a lot of different people at parties.\\
\noindent Q56: Am not really interested in others.\\
\noindent Q57: Like order.\\
\noindent Q58: Change my mood a lot.\\
\noindent Q59: Am quick to understand things.\\
\noindent Q60: Don’t like to draw attention to myself.\\
\noindent Q61: Take time out for others.\\
\noindent Q62: Shirk my duties.\\
\noindent Q63: Have frequent mood swings.\\
\noindent Q64: Use difficult words.\\
\noindent Q65: Don’t mind being the center of attention.\\
\noindent Q66: Feel others’ emotions.\\
\noindent Q67: Follow a schedule.\\
\noindent Q68: Get irritated easily.\\
\noindent Q69: Spend time reflecting on things.\\
\noindent Q70: Am quiet around strangers.\\
\noindent Q71: Make people feel at ease.\\
\noindent Q72: Am exacting in my work.\\
\noindent Q73: Often feel blue.\\
\noindent Q74: Am full of ideas.\\
\noindent Q75: What is your MBTI personality type? (Optional)  
\\

\section{Interview Design}
\label{Interview-protocol}
The interview includes question sections of \textbf{demographic info, detailed personal background, and online shopping habits and preferences}. The interviews encourages participants to elaborate on their personas through open responses, thus facilitating a more nuanced understanding of their individual experiences and decision-making processes. 
For example, in the personal narrative section, participants were encouraged to describe a typical day and share how they perceive themselves. Similarly, in the online shopping-related section, they were asked to describe a recent purchase session on Amazon, providing concrete examples of their shopping habits for downstream models' understanding and simulation of consumer behavior. This includes pre-purchase research activities, in-session shopping behaviors, engagement with review content, and attitude on advertisement. 

\subsection{Interview Protocal}
\noindent\textbf{Demographics}\\
\textbf{Introduction:}  
Can you tell me a bit about yourself? What kind of work do you do? Where do you live? Do you live alone or with family?  

\noindent\textbf{Personal Background}\\
\textbf{Daily Life:}  
You mentioned your work/study. What does a typical day look like for you?  

\noindent\textbf{Work Activities:}  
Can you tell me more about [job/study]? What are your main responsibilities or daily tasks?  

\noindent\textbf{After-Work Activities:}  
What do you usually do after work?  

\noindent\textbf{Self-Perception:}  
How would you describe yourself?  

\noindent\textbf{Online Shopping Preferences}\\
\textbf{Recent Purchase:}  
Can you recall a recent Amazon purchase? What was the reason for shopping? What were you looking for? How did you find and decide on the product?  

\noindent\textbf{Pre-Shopping Activities:}  
Did you research before making the purchase? 

\noindent\textbf{During-shopping Shopping:}  
How long did it take to decide? How many products did you compare before choosing?  

\noindent\textbf{Decision Factors:}  
What mattered most in your choice? Do your priorities change based on category (e.g., style for clothing, brand for electronics, price for essentials)?  

\noindent\textbf{Reviews:}  
Did you read any reviews before purchasing? What information were you looking for? If not, why?  

\noindent\textbf{Advertisements:}  
Do you notice sponsored products? How do ads influence your decisions?

\section{Web Parser Design}
\label{recipe-design}

This section introduces the parser designed to process and simplify web pages, enabling downstream LLMs to better understand and interact with the page content. The parser is guided by a framework called a recipe, which consists of a set of JSON-based rules tailored to specific web pages. These recipes are created through a combination of manual rule-writing and automated assistance from GPT.

Each recipe uses CSS selectors to identify and extract key HTML elements that are important for user interaction (e.g., search boxes, filter options) or containing important semantic information (e.g., product prices, availability status). 

We designed recipes for several common page types, such as search result pages, product detail pages, and checkout pages. When parsing a page, the parser extracts these key elements and annotates them with a unique semantic ID. For instance, the ID \texttt{refinements.colors.red} denotes an element that filters results to red-colored items when clicked. These semantic IDs help the model understand both the structure and the function of the elements. Clickable elements are further annotated with visual markers to inform the downstream model of their interactivity.

To simplify the HTML, the parser removes all irrelevant content and styling, retaining only the extracted key elements. This results in a clean, minimal HTML structure that is easier for the model to analyze and reason over.

\section{Rationale Prompt Design}
\label{reasoning-design}
The plugin employs context-aware pop-up questions to collect user rationale across different interaction types. The question design follows a hierarchical structure based on event types and specific interaction targets.
\\
\\
\textbf{Click Events}
\begin{itemize}
  \item \textit{Click on subscription setup button}: You clicked on the set up now button. Can you tell us why you subscribed to this product?
  \item \textit{Click on buy now button}: You clicked on the "buy now" button. Why did you do that?
  \item \textit{Click on add to cart button}: You clicked on the "add to cart" button. Why did you decide to add this product to your cart?
  \item \textit{Click on search button}: You clicked on the "search" button. Why did you make this search?
  \item \textit{Click on filters}: You clicked on this filter. Why did you use this filter?
  \item \textit{Click on product options}: You clicked on this product option. Why did you click this product option?
  \item \textit{Click on checkout button}: You clicked on the "checkout" button. What made you decide to checkout?
  \item \textit{Click on decrease quantity button}: You clicked on the “decrease quantity” button. Why did you click this button?
  \item \textit{Click on increase quantity button}: You clicked on the “increase quantity” button. Why did you click this button?
  \item \textit{Click on product list}: You clicked on this product. Why did you click on this product?
  \item \textit{Click on other area}: We noticed that you just had a click action. Why did you do that?
\end{itemize}

\textbf{Scroll Events}
\begin{itemize}
  \item \textit{Page scrolling}: We saw that you scrolled up/down this page. What are you looking for?
\end{itemize}

\textbf{Navigation Events}
\begin{itemize}
  \item \textit{Back / forward navigation}: Why did you decide to return to this page?
\end{itemize}

\textbf{Tab Switch Events}
\begin{itemize}
  \item \textit{Tab activation}: Why did you leave and come back to this tab?
\end{itemize}

\section{Segmentation Interval Threshold Justification}
\label{intervalDistribution}

To determine the time interval threshold for session segmentation, we analyzed the distribution of inter-action time gaps from the first week of user activity. From observation, approximately 98\% of intervals were shorter than 4 minutes, and 99\% were shorter than 78 minutes. Choosing the 99th percentile (78 minutes) as the threshold balances session granularity, avoiding both excessive fragmentation and excessively long sessions that mix unrelated behaviors.

In addition, behavioral analysis showed that users typically interact with the site for 50 to 150 actions before reaching a natural session boundary, such as a purchase or exit, which means around every 50 to 150 interactions, there is probably a session termination or purchase signal. This observation supports the chosen 99\% threshold as both statistically and behaviorally grounded.

\section{Click Type Description}
\label{app:click_type_explanation}

This section provides detailed descriptions for the click type categories used in the OPeRA dataset, as presented in Table~\ref{tab:click_statistics}.
The following categories were established through empirical analysis of user click action distributions.

\noindent\textbf{review} (20.8\%): Clicks on review-related elements, including review images, star ratings, review filters, etc. 

\noindent\textbf{search} (15.1\%): Clicks on search button or search box.

\noindent\textbf{product\_option} (13.9\%): Clicks on product options such as size selectors, color options etc.

\noindent\textbf{product\_link} (10.6\%): Clicks on product images, product titles, or product links that navigate users to product detail pages.

\noindent\textbf{other} (8.9\%): Miscellaneous clicks that do not fall into the above predefined categories.

\noindent\textbf{purchase} (6.4\%): Clicks on purchase-intention elements including ``Add to Cart'', ``Buy Now'', ``Subscribe'', and ``Checkout''.

\noindent\textbf{nav\_bar} (5.6\%): Clicks on navigation bar elements such as category menus, amazon logo etc.

\noindent\textbf{page\_related} (3.9\%): Clicks on pagination controls, carousel navigation buttons that control page content display.

\noindent\textbf{quantity} (3.8\%): Clicks on quantity adjustment controls including increase/decrease buttons and item deletion buttons.

\noindent\textbf{suggested\_term} (3.6\%): Clicks on search suggestions.

\noindent\textbf{cart\_side\_bar} (2.9\%): Clicks on elements in shopping cart sidebar.

\noindent\textbf{cart\_page\_select} (2.8\%): Clicks on selection elements such as item checkbox in the cart page.

\noindent\textbf{filter} (1.8\%): Clicks on filtering elements including price filters, brand filters, rating filters, and other product refinement controls. 





\section{Next Action Prediction Experiment Results}
\label{app:action_prediction_full_result}
Table~\ref{tab:action_predict_filtered} shows the full results of next action prediction task.
\begin{table*}[t]
\centering
\begin{booktabs}{
  colspec={lcccccc},
  cells={font=},
  row{1}={font=\bfseries\small},
}
\toprule
\textbf{Model} &
\makecell{Next Action Gen.\\(Accuracy)} &
\makecell{Action Type\\(Weighted F1)} & 
\makecell{Action Type\\(Macro F1)} & 
\makecell{Click Type\\(Weighted F1)} &
\makecell{Outcome\\(Accuracy)} & 
\makecell{Outcome\\(Weighted F1)} \\
\midrule
GPT-4.1 & 21.51 & \textbf{85.04} &\textbf{ 48.78} & \textbf{44.47} & 38.89 & 47.54 \\
\hspace{0.4em}w/o persona & \textbf{22.06} & 82.32 & 45.55 & 43.45 & 55.55 & \textbf{58.47} \\
\hspace{0.8em}w/o rationale & 21.28 & 83.13 & 34.93 & 42.63 & 53.33 & 51.17 \\
\midrule
DeepSeek-R1 & 14.75 & 81.99 & 27.37 & 35.12 & 51.11 & 46.36 \\
\hspace{0.4em}w/o persona & 15.52 & 81.72 & 27.43 & 33.86 & \textbf{56.67} & 48.86 \\
\hspace{0.8em}w/o rationale & 15.74 & 81.66 & 27.16 & 32.65 & 53.33 & 47.92 \\
\midrule
Claude-3.7 & 10.75 & 83.41 & 31.58 & 27.27 & 52.22 & 43.52 \\
\hspace{0.4em}w/o persona & 10.75 & 82.28 & 25.33 & 22.76 & 50.00 & 43.10 \\
\hspace{0.8em}w/o rationale & 10.08 & 81.08 & 26.06 & 20.29 & 47.78 & 43.10 \\
\midrule
Llama-3.3 & 8.31 & 80.69 & 24.29 & 19.99 & 34.44 & 36.64 \\
\hspace{0.4em}w/o persona & 8.31 & 78.59 & 23.69 & 18.59 & 28.89 & 33.21 \\
\hspace{0.8em}w/o rationale & 8.76 & 80.23 & 23.60 & 19.22 & 31.11 & 34.19 \\
\bottomrule
\end{booktabs}
\caption{
Evaluation of actions in next action prediction task. ``Claude-3.7'': Claude-3.7-Sonnet, ``Llama-3.3'': Llama-3.3-70B-Instruct. All metrics are reported as percentages (\%). Instance size $n=902$.
}
\label{tab:action_predict_filtered}
\end{table*}

\section{Experiment Prompt Design}
Below are the two prompts for action prediction task and joint rationale and action generation task:
\label{prompt}
\begin{lstlisting}[style=promptstyle]
PROMPT_FOR_ACTION_PREDICTION = """
<IMPORTANT>
Your task is to predict the immediate next action of a shopper.
You need to pretend that you are a real user shopping on amazon.com.
The history action, rationale, context and the user persona will be provided to you.
Ensure your prediction follows natural behavior sequences (e.g., users may click a search box before typing, type a query before clicking search)
</IMPORTANT>

# Action Space

An action is represented in JSON format, and there are four primary types of actions:

#### 1. `input`:
Type text into an input field.
{
    "type": "input",
    "name": "input_name",
    "text": "input_text"
}

#### 2. `click`:
Click on a button or clickable element identified by `name`.
{
    "type": "click",
    "name": "clickable_name",
}

#### 3. `terminate`:
When you are unsatisfied with the current search result and you don't want to buy anything, use `terminate` to indicate that you want to close the browser window and terminate the task.
{
    "type": "terminate"
}

# Rationale
Rationale is the reason why the user takes the action. Some of the rationale is provided to you.

# Context
Your context will be the HTML of the amazon page you are looking at. Some interactable elements will be added a unique "name" attribute, which you can use to identify the element to interact with (click or input).

# Persona
The user persona reflects the user's demographics, personality, and shopping preference. First identify which aspects of the persona might be relevant to the current shopping context, then consider them only if they naturally align with the ongoing shopping journey. DO NOT RELY ON IT.

# Output Format
You need to predict the next action. Your output should follow a strict JSON format:
{
    "type": "<type>",
    ...
}

<IMPORTANT>
OUTPUT A SINGLE JSON OBJECT, NOTHING ELSE.
</IMPORTANT>
"""


"""

\end{lstlisting}

\end{document}